\documentclass{article}

\usepackage[final]{corl_2026} 

\usepackage{amsmath}
\usepackage{amssymb}
\usepackage{graphicx}
\usepackage{siunitx}
\usepackage{booktabs}
\usepackage{enumitem}
\usepackage{hyperref}

\title{Benchmarking Real-World Robotic Manipulation through Stochastic Challenges}

\author{
  Elias Gl\"a\ss er$^{1,\dagger}$,
  G\"ulhan \c{S}ikaro\u{g}lu$^{1,2,\dagger}$,
  Pavlo Kupyn$^{1}$,
  Marc Pollefeys$^{3,4}$,\\
  \textbf{Zuria Bauer}$^{3}$\textbf{,}
  \textbf{Hermann Blum}$^{1,2,\dagger}$\\[0.5em]
  $^{1}$ University of Bonn
  $^{2}$ Lamarr Institute
  $^{3}$ ETH Zürich
  $^{4}$ Microsoft
}
 
\begin{document}
\raggedbottom
\maketitle
\begingroup
\renewcommand{\thefootnote}{\dag}
\footnotetext[1]{
  Correspondence to:
  \texttt{glaesser.elias@gmail.com},
  \texttt{gsikaroglu@gmail.com},
  \texttt{blumh@uni-bonn.de}
}
\endgroup


\begin{abstract}
A challenging bottleneck in manipulation research is reproducibility and comparability of methods in real-world environments. Since manipulation is an interactive task, it cannot be evaluated on a static dataset of observations. Benchmarking in simulation is a common remedy, but cannot cover the whole complexity of real-world tests. This paper proposes a new paradigm of \emph{stochastic challenges}. Like the well-known YCB objects, it is based on reproducible real-world scenes and therefore allows system-level innovation and comparability between different platforms. Due to their stochastic nature, our challenges however, mimic the rigor and surprise of hidden test sets. 
To this end, we show how different robots in different labs can be compared based on their success rate over different random task samples, if these samples come from the same distribution. We propose stochastic challenges as a general paradigm for robot benchmarking and implement this paradigm in a novel benchmark for mobile manipulation, on which we compare two robots tested in two different labs.
\end{abstract}
\begin{center}
    \small
    \textbf{Project Website:}
    \href{https://rpl-bonn.github.io/stochastic-challenges/}{\textbf{\textcolor{blue}{Stochastic Challenges}}}
    \quad | \quad
    \textbf{Benchmark:}
    \href{https://klxbench.rpl.uni-bonn.de/}{\textbf{\textcolor{blue}{KLX Benchmark}}}
\end{center}


\section{Introduction}

Reliable comparability between robotic manipulation methods is a common bottleneck in current robotics research. Manipulation is an inherently interactive task: the same model can produce different outcomes on the same physical setup, and different model outputs can lead to the same outcome, so evaluation on static datasets cannot capture method quality. Existing alternatives each carry limitations. Centralized competitions such as RoboCup require physically transporting robots to a single site. Simulation benchmarks \cite{Pum24} provide scale and repeatability but exhibit physical realism gaps. Real-world benchmarks built around fixed object sets \cite{Cal15} have driven important progress, but a single hand-picked setup risks overfitting.

\begin{figure}[t]
    \centering
    \includegraphics[width=.8\linewidth]{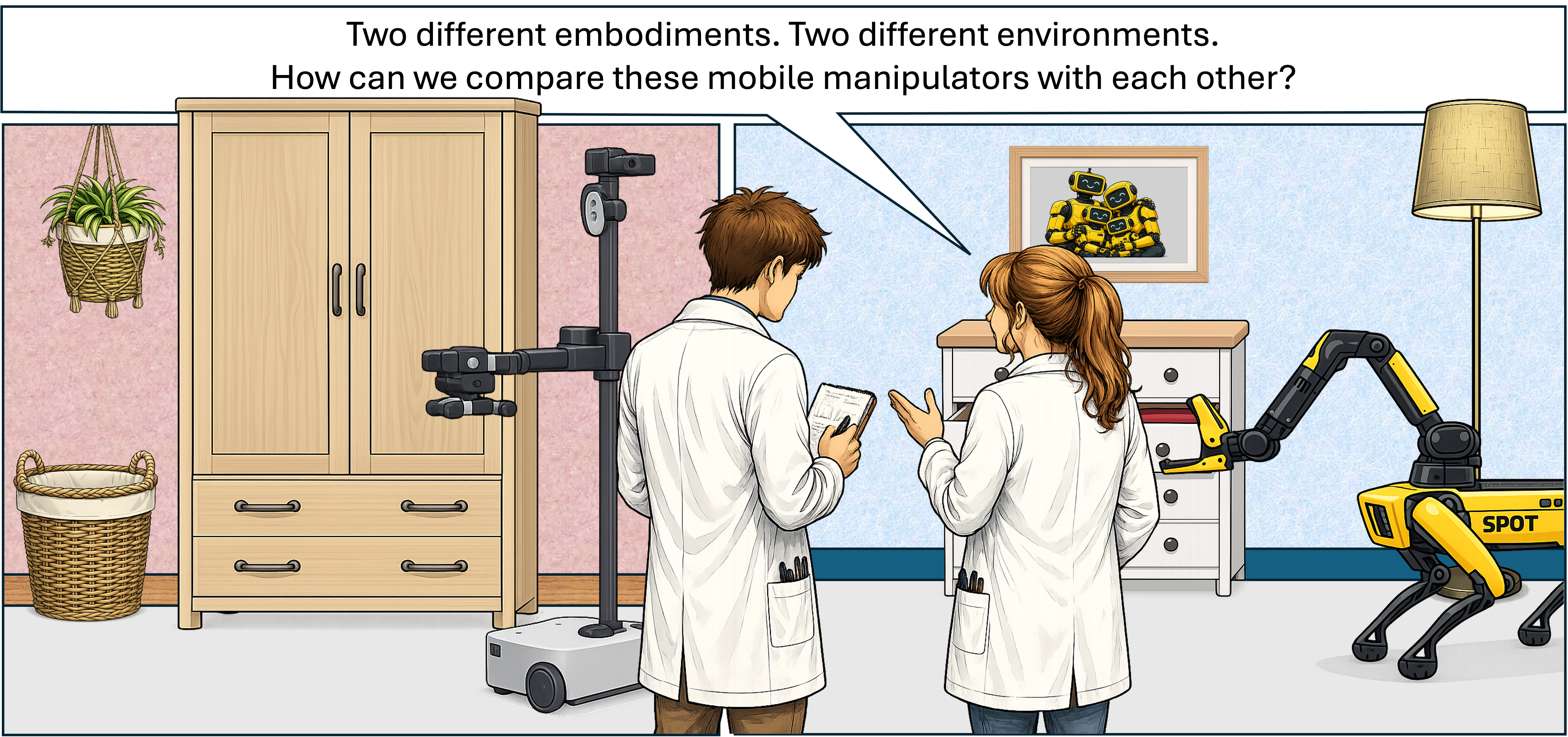}
    \caption{\textbf{Stochastic Challenges} target several desired aspects of robotic benchmarking: (i) system-level innovation, i.e. not restricting evaluation to specific hardware; (ii) local evaluation, avoiding the effort to transport systems and reducing the chance of errors when trying to reproduce methods; (iii) reduced overfitting, testing true generalization instead of pre-known tasks.}
    \label{fig:teaser}
\end{figure}

Recent efforts have identified the same problem and proposed real-world distributed benchmarks \cite{Atr25, Che25b}. However, both rely on 
standardizing either the robot embodiment \citep{Atr25} or the task \cite{Che25b}, and both constrain themselves to tabletop scenarios. This prevents precisely the comparisons that matter as manipulation systems scale to deployment: comparing different platforms - humanoids, wheeled mobile manipulators, quadrupeds, on real tasks that require both mobility and interaction with real scenes.

We address this gap with \emph{stochastic challenges}, a benchmarking framework that standardizes a \emph{distribution} of scenes and tasks rather than a fixed test set. This extends the idea of YCB \cite{Cal15} from a fixed object set to a sampled distribution: any lab can reconstruct any specific scene from a seed, but participants see only a small fraction of all possible scenes and do not know the correct response to a challenge in advance, similar to the test set of a dataset benchmark. This allows evaluation instances to be sampled only at test time, unlike a fixed paired test set. A system must therefore be prepared for the broader task distribution even though only a small subset of configurations is evaluated in practice. This reduces the opportunity to overfit to a known physical test set. We hypothesize that, if tasks and scenes are sampled from the same distribution, sufficient samples allow methods to be compared at a chosen statistical confidence level independent of the robot platform, software stack, or policy used.

We instantiate this paradigm in a concrete benchmark built on IKEA KALLAX shelves: a widely available household furniture system that provides standardized geometry, modular inserts (doors, drawers, open shelves), and a natural set of interaction primitives, such as opening, retrieving, placing, and searching. A seeded randomizer generates configurations and object placements, each identified by a seed and exported as a spec from which any partner lab can reconstruct the same physical setup. We demonstrate the paradigm by comparing two robots tested in two different labs and analyzing how many randomized samples are required to draw confident conclusions.

In summary, our main contributions are:
\begin{itemize}[nosep]
    \item The \emph{stochastic challenges} framework for real-world manipulation benchmarking, together with a variance-decomposition framework that quantifies how many randomized samples are needed 
    for statistically grounded cross-method comparisons.
    \item The \emph{KALLAX benchmark} -- 
    comprising a seeded configuration randomizer, a task distribution, and a web platform for cross-lab distribution and reproducibility.
    \item A study comparing two robotic systems with different embodiments in two different labs on the benchmark, validating the statistical framework and characterizing the sample complexity required to separate methods.
\end{itemize}


\section{Related Work}
\label{sec:relatedworks}

\subsection{Simulation benchmarks}
Because simulation offers repeatability at low cost, many manipulation benchmarks are built in simulation. RLBench~\cite{Jam20} established a large-scale standardized suite spanning many manipulators and learning paradigms, while later benchmarks probe different aspects of generalization. Colosseum~\cite{Pum24} studies robustness under visual and physical perturbations, \cite{Gar25} evaluate LLM-guided 3D policies for category-level generalization across articulation and placement, and \cite{Zhao25, Zha25c} examine vision-language reasoning for low-level and spatially grounded tabletop tasks. CALVIN~\cite{calvin}, LIBERO~\cite{libero}, and ARNOLD~\cite{arnold} further evaluate generalization across predefined tasks, environments, objects, or goal states, while RoboCasa~\cite{Nas24} and BEHAVIOR-1K~\cite{Li22} broaden evaluation toward large-scale realistic household activities and environments beyond the tabletop settings. 

These benchmarks therefore already contain stochasticity or controlled variation, but the source of randomness is typically defined within a predetermined simulator, task suite or test split. In contrast, \emph{stochastic challenges} sample the evaluation examples themselves: complete physical scenes and tasks are drawn at evaluation time from a specified distribution. This distinction is important in real-world manipulation, where perception errors, contact dynamics, execution failures, and environment-specific variation introduce uncertainty that cannot be fully reproduced in simulation.

\subsection{Real-world manipulation benchmarks}
Real-world benchmarks aim to provide comparability across physical setups by standardizing objects, layouts, tasks, or hardware. Some focus on reproducible tabletop manipulation through common object sets and physical arrangements, including YCB, GRASPA, Box-and-Blocks, and SceneReplica~\cite{Cal15, graspa, Morgan20, scenereplica}. However, repeatedly evaluating on the same objects and scenes can encourage overfitting and constrain the range of tasks. Centralized competitions such as RoboCup instead enforce comparability by gathering robots at a single site, which limits participation to those able to physically attend. Others standardize broader task protocols or hardware platforms, such as RB2, NIST Assembly, OCRTOC, TOTO, and Real Robot Challenge~\cite{RB2,NIST, OCRTOC, TOTO, Bauer2021Real}. More recent benchmarks extend this direction toward complex manipulation or distributed evaluation: 
FurnitureBench~\cite{Heo25} evaluates RL and IL models on miniature 3D-printed furniture, where miniature-scale interactions may not fully reproduce full-scale contact dynamics, and FMB~\cite{Luo25} benchmarks multi-object manipulation with engineered blocks and multiple sensing modalities, though with tasks centered on functional manipulation rather than semantically rich household interaction. ManipulationNet and RoboArena~\cite{Che25b, Atr25} extend evaluation across platforms or institutions, but they remain confined to tabletop manipulation. RoboDojo~\cite{robodojo} further combines simulation and real-world evaluation across a broader set of manipulation tasks.

Despite this breadth, comparability is typically obtained through matched objects, layouts, tasks, or embodiments, often under tabletop or fixed-base settings. Such \emph{paired} evaluation provides strong experimental control, but requires systems to be evaluated on matched physical instances, which in fixed benchmarks may already be known during method development and experiment preparation.

Our framework instead standardizes the \emph{distribution} from which real-world scenes and tasks are sampled. Robots may therefore execute different samples drawn from the same distribution, enabling unpaired comparison across laboratories and embodiments without requiring identical hardware, software or scene. Seeded generation supports matched configurations when desired, while independent seeds enable unpaired evaluation. Moreover, whereas prior physical benchmarks commonly report aggregate success rates, standard deviations, or ranking statistics, our variance-based framework (Section~\ref{sec:method}) determines whether differences observed over finite stochastic trials are statistically distinguishable.
	

\section{Method: Stochastic Challenges}
\label{sec:method}

\begin{figure}[t]
    \centering
    \includegraphics[width=\linewidth]{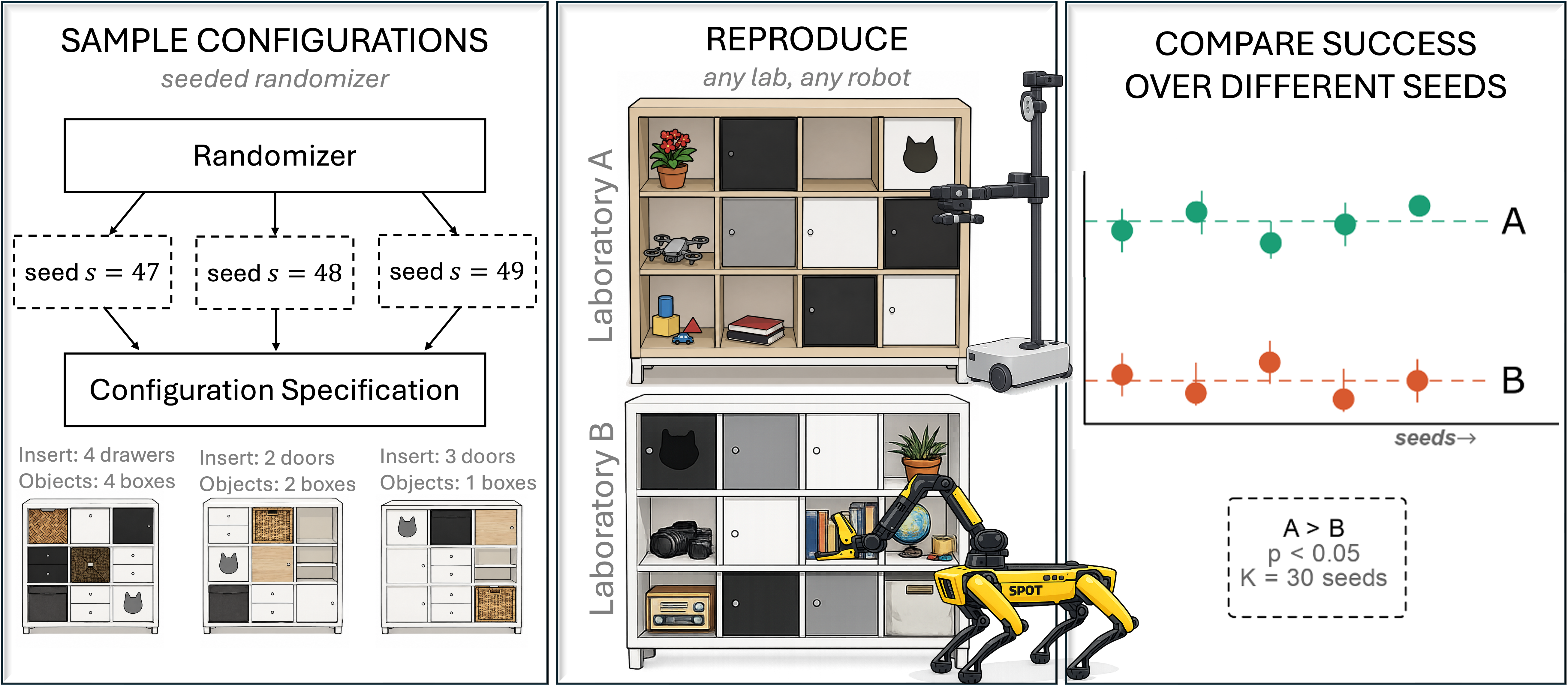}
    \caption{Concept overview. Seeded randomization generates reproducible KALLAX manipulation scenes; robots across labs (Stretch~3 and Spot) execute sampled configurations, with outcomes statistically aggregated for robust comparison.}
    \label{fig:concept}
\end{figure}

Figure~\ref{fig:concept} summarizes the proposed stochastic-challenge paradigm. Rather than evaluating methods on fixed setups, we evaluate them on random configurations sampled from a defined distribution. Aggregating binary success outcomes across sampled configurations estimates expected performance over the full distribution.
Sampling configurations at evaluation time gives the benchmark its rigor: participants cannot tune to known setups, while experiments remain independently reproducible across labs without shared hardware constraints. However, because results depend on the sampled configurations and preventing overfitting requires new samples across experiments, a key question arises: how many configurations and trials are needed to draw statistically significant conclusions? The remainder of this section develops the framework that answers this. 

The framework supports three comparison regimes: independently sampled configurations across systems, fully shared configurations, and mixed designs in which only a fraction of seeds is shared. These correspond to different trade-offs between generalization to unseen configurations and statistical efficiency, developed in Sections~\ref{sec:method-setup}--\ref{sec:assumptions}.

\subsection{Setup and benchmark score}
\label{sec:method-setup}

We evaluate a robot $R$ on scenes drawn from the randomized distribution
created by the seeded configuration generator (Section~\ref{sec:kallax}).
A \emph{seed} $s$ therefore fixes a complete scene layout / configuration. We further assume that $p(s)$ is uniform, i.e. each scene has the same likelihood of being sampled.
For a given scene, we define a set of $M_s$ tasks that a robot may attempt.
Each task $i$ may be attempted several
times, and every attempt $j$ yields a binary outcome
$Y_{Rsij}\in\{0,1\}$ indicating task success. As is commonly done in robotic benchmarking, we model attempts for each task as independent Bernoulli trials with an unknown
per-task success probability $p_{Rsi}$; the elementary attempt- and
task-level moments are standard and are deferred to
Appendix~\ref{app:estimators}. In practice, we approximate this assumption by resetting the robot software and task state between repeated attempts.

The quantity we ultimately report is the robot's expected success over
the distribution of scenes and tasks. Denoting $\theta_{Rs} = \frac{1}{M_s}\sum_{i=1}^{M_s} p_{Rsi}$ as the robot's
true mean success on seed $s$, the benchmark objective is $\mu_R \;=\; \mathbb{E}_{s}\left[\theta_{Rs}\right]$,
the expectation taken over seeds $s$ drawn from the benchmark
distribution; Measuring $\mu_R$ rather than performance over a
manually chosen scene or set of scenes is what makes the benchmark a property of the
\emph{distribution} rather than of specific setups.

Because $p_{Rsi}$ and $\mu_R$ are unobservable, we estimate them by a
three-level average -- over repeated attempts per task, tasks within a
scene, and over scenes:
\[
    \hat p_{Rsi} = \frac{1}{J}\sum_{j=1}^{J} Y_{Rsij},
    \qquad
    \hat\theta_{Rs} = \frac{1}{M_s}\sum_{i=1}^{M_s} \hat p_{Rsi},
    \qquad
    \hat\mu_R = \frac{1}{K}\sum_{s=1}^{K} \hat\theta_{Rs},
\]
where $J$ is the number of repeated attempts per task and $K$ the
number of sampled seeds / scenes. Averaging to a seed-level score
$\hat\theta_{Rs}$ before averaging over seeds ensures that scenes with
more possible tasks, or tasks that happened to be attempted more often, do
not dominate the benchmark mean. Under the sampling assumptions of
Section~\ref{sec:assumptions}, $\hat\mu_R$ is unbiased for $\mu_R$
(Appendix~\ref{app:estimators}).

The two choices the experimenter controls are $J$ and $K$: how
often each task is attempted, and how many randomized seeds are
evaluated. The remainder of this section asks how the uncertainty of
$\hat\mu_R$ depends on these two quantities, and how they should be
chosen.

\subsection{Two sources of variance}
\label{sec:method-variance}

The uncertainty in $\hat\mu_R$ has two qualitatively different origins. The
first is \emph{trial noise}: even on a single fixed configuration a robot
succeeds on some attempts and fails on others, so a task's success
rate is estimated from only $J$ Bernoulli trials. The second is
\emph{configuration difficulty}: randomized scenes are likely to differ in
how difficult they are. Therefore, we cannot control for the difficulty at benchmark level and the per-seed
performance of a given robot $\theta_{Rs}$ varies from seed to seed. The spread
of per-seed performance across the distribution $\sigma^2_{\mathrm{seed},R} = \mathrm{Var}_S(\theta_{Rs})$
then measures how strongly $\hat\mu_R$ depends on which
scenes happen to be sampled.

Treating a single seed-level score $\hat\theta_{Rs}$ as random over both
the sampled seed and the trial noise on it, the law of total variance
separates these two contributions exactly
(Appendix~\ref{app:variance}). Under approximately equal $J$ trials per
task, the result is a \emph{planning variance}
\[
    V_R(J)
    \;=\;
    \mathrm{Var}(\hat\theta_{Rs})
    \;=\;
    \underbrace{\sigma^2_{\mathrm{seed},R}}_{\text{between configurations}}
    \;+\;
    \underbrace{\frac{B_R}{J}}_{\text{within configuration}},
\]
where $B_R$ aggregates the per-task Bernoulli variance
$p_{Rsi}(1-p_{Rsi})$, reduced by averaging over the $M_s$ tasks of a
seed and over the seed distribution; its exact form is given in
Appendix~\ref{app:variance}. 
Averaging over $K$ independently sampled seeds gives the variance of the
reported benchmark mean,
\[
    \mathrm{Var}(\hat\mu_R)
    \;=\;
    \frac{V_R(J)}{K}
    \;=\;
    \frac{\sigma^2_{\mathrm{seed},R} + B_R/J}{K}.
\]
This expression makes the allocation problem explicit. Both terms are
divided by the number of seeds $K$, but only the second is further
reduced by the number of attempts $J$. In particular, as $J \to \infty$
the benchmark variance does not vanish but approaches a floor
$\sigma^2_{\mathrm{seed},R}/K$, set entirely by the number of evaluated
configurations: no amount of per-configuration repetition can beat this
floor, and it can be lowered only by sampling more seeds.  The framework, however, does not require configurations to have equal difficulty. Greater variation across seeds instead appears as a larger between-configuration variance and consequently increases sample complexity.


\subsection{Sample complexity}
\label{sec:method-samplesize}

The variance expression of Section~\ref{sec:method-variance} turns
directly into experiment-planning rules. For a given robot, a
two-sided $100 \cdot(1-\alpha)\,\%$ confidence interval for $\mu_R$ has
half-width $h = z_{1-\alpha/2}\sqrt{V_R(J)/K}$, where $z_q$ denotes the
$q$-quantile of the standard normal distribution, so reporting the
benchmark score to a target precision $\varepsilon$ requires
$K \ge z_{1-\alpha/2}^2\,V_R(J)/\varepsilon^2$ seeds.

A benchmark's primary purpose, however, is comparison. For two robots
$A$ and $B$ we estimate their performance gap $\hat\Delta = \hat\mu_A - \hat\mu_B$;
when each robot is evaluated on its own set of independently sampled seeds, the
two estimates are independent, so the comparison variance is
$\mathrm{Var}(\hat\Delta) = 2V(J)/K$ (assuming comparable planning
variance $V(J)$ for both). Under the normal approximation, detecting a
true gap of size $\delta$ at significance level $\alpha$ and power
$1-\beta$ requires
\begin{equation}
\label{eq:samplesize}
    K
    \;\ge\;
    \frac{
        2\left(z_{1-\alpha/2} + z_{1-\beta}\right)^2 V(J)
    }{\delta^2}
    \;=\;
    \frac{
        2\left(z_{1-\alpha/2} + z_{1-\beta}\right)^2
        \left(\sigma^2_{\mathrm{seed}} + B/J\right)
    }{\delta^2}
\end{equation}
seeds per robot.
The significance level $\alpha$ is the probability of a Type 1 error: concluding a significant advantage for one robot that does not truly exist. $\beta$ is the probability of a Type 2 error: Not being able to find an advantage through the benchmark that actually exists.

Equation~\eqref{eq:samplesize} exposes the cost structure of the
benchmark. The required number of seeds grows with the
configuration-difficulty spread $\sigma^2_{\mathrm{seed}}$ and the trial
noise $B/J$, and falls quadratically as the gap $\delta$ to be resolved
grows: small differences between strong systems are expensive to
certify. The trial-noise contribution $B/J$ can be reduced
by raising $J$, but the $\sigma^2_{\mathrm{seed}}$ contribution cannot:
it is fixed by the distribution and, under independent seeds, imposes a
hard lower bound $K \ge 2(z_{1-\alpha/2}+z_{1-\beta})^2
\sigma^2_{\mathrm{seed}}/\delta^2$ that persists even as $J \to \infty$.
Because this dominant term arises purely from testing the two robots on
\emph{different} sampled configurations, it can be eliminated by testing
them on the \emph{same} ones; we develop this in
Section~\ref{sec:method-comparison}. Finally, since
$\sigma^2_{\mathrm{seed}}$ and $B$ are not known in advance and must be
estimated from a pilot set of seeds, \eqref{eq:samplesize} should be read
as a planning estimate rather than an exact requirement
(Section~\ref{sec:assumptions}).


\subsection{Comparing robots across labs}
\label{sec:method-comparison}

The dominant cost in \eqref{eq:samplesize} is the between-configuration
term $\sigma^2_{\mathrm{seed}}$, and it arises only because the two
robots were evaluated on \emph{different} sampled configurations. If
instead both robots are run on the \emph{same} seeds, the difficulty of
each configuration acts on both equally and cancels in their per-seed
difference. Writing the per-seed gap
$D_s = \hat\theta_{As} - \hat\theta_{Bs}$ under a paired random-effects
model in which a common seed-difficulty effect enters both robots'
scores (Appendix~\ref{app:comparison}), that shared effect subtracts
out, leaving only the robots' relative skill and the residual trial
noise.

More generally, let $p \in [0,1]$ be the fraction of seeds shared by
both robots. The comparison variance then interpolates between the
independent and fully paired designs:
\begin{equation}
\label{eq:mixedvar}
    \mathrm{Var}(\hat\Delta)
    \;=\;
    \frac{2\left[(1-p)\,\sigma^2_{\mathrm{seed}} + B/J\right]}{K}.
\end{equation}
At $p=0$ this recovers the independent-seed case of
Section~\ref{sec:method-samplesize}; at $p=1$ the between-configuration
term vanishes entirely, leaving $\mathrm{Var}(\hat\Delta) = 2B/J\,/\,K$
-- a fully paired design pays only the trial-noise price.

Substituting \eqref{eq:mixedvar} into the power requirement gives the
number of seeds needed to detect a gap $\delta$ under partial sharing:
\begin{equation}
\label{eq:mixedss}
    K
    \;\ge\;
    \frac{2\left(z_{1-\alpha/2}+z_{1-\beta}\right)^2
          \left[(1-p)\,\sigma^2_{\mathrm{seed}} + B/J\right]}{\delta^2}.
\end{equation}
Fully paired evaluation ($p=1$) is statistically efficient because both systems encounter the same configurations. However, when the complete physical test set is known in advance, repeated evaluation may encourage adaptation to those particular objects and scenes rather than measuring generalization over the broader challenge distribution. At the other extreme, independent sampling ($p=0$) better evaluates generalization to unseen configurations but may require more trials. Intermediate values $0<p<1$ provide a practical compromise between these objectives, with the appropriate choice depending largely on the cost of reconstructing physical scenes.

\subsection{Assumptions}\label{sec:assumptions}

We assume independent seed sampling from the same benchmark distribution and require a central service (e.g. web service) to prevent selective reporting. We model repeated attempts as independent Bernoulli trials and approximate this assumption by resetting the task and software state. We capture variations in seed difficulty through $\sigma^2_{\mathrm{seed},R}$ and assume a common difficulty effect only for paired comparisons. Finally, we use a normal approximation of aggregated seed-level scores for sample-complexity analysis.

\section{The KALLAX Benchmark}\label{sec:kallax}
Based on the benchmarking methodology in Section~\ref{sec:method}, we propose a mobile manipulation benchmark for robotic furniture interaction.
We focus on an easily reproducible physical environment (the IKEA KALLAX shelf) and randomized configuration generation to obtain a large number of comparable real-world trials. 

\textbf{Randomizer} We randomly configure the $3\times4$ KALLAX grid by first uniformly selecting a specified number of empty positions without replacement. In our experiments, an empty fraction of $0.5$ results in six empty positions. 
Each remaining position is assigned an insert from a configurable pool of up to 82 KALLAX-compatible options, including doors and drawers. The evaluated configurations contain approximately $2.5$ times as many door as drawer inserts, with each drawer insert comprising two drawers. Object-search tasks additionally sample from 79 YCB objects. Each configuration has a unique seed and is exported in machine- and human-readable form with a rendered visualization. 

\textbf{Tasks and success criteria}
On top of the randomized setups we define manipulation tasks that reflect household interactions and are executable on a mobile manipulator, for example: (1) Open: open a specified door/drawer to a target state, (2) Pick: retrieve a specified object from a specified shelf/drawer, (3) Place/Move: move an object to a target cubby or shelf position. Objects to move/ search for may be randomized as well as the shelf’s general configuration. This creates many trials with controlled variability while keeping tasks understandable and close to real-world use cases.


\textbf{Web interface and cross-lab reproducibility}
We provide a publicly accessible website that
(i) generates configurations (with seeds),
(ii) presents assembly instructions and object lists,
and (iii) exports task instances and records reported outcomes.
The generator is available through the
\href{https://klxbench.rpl.uni-bonn.de/}{KLX Benchmark}%
\footnote{\url{https://klxbench.rpl.uni-bonn.de/}}.

A partner lab may reproduce a configuration by following the same specification, execute the same task set, and upload results. This supports direct replication studies (same configuration) and generalization studies (different samples from the same distribution). The platform records generated configurations and reported outcomes to support transparency and discourage selective reporting.




\textbf{Reliable success labeling} Some failures are ambiguous (partial opening, object grasped but dropped, slight miss on the knob). Clearly defined success criteria per task and failure type are essential. Long-term automated verification (e.g., VLM-based video checks) is promising, but for this project, we validate labels through careful logging and human review.


\section{Experiments}
\label{sec:experiments}

\begin{figure}[t]
    \centering
    \includegraphics[width=0.7\textwidth]{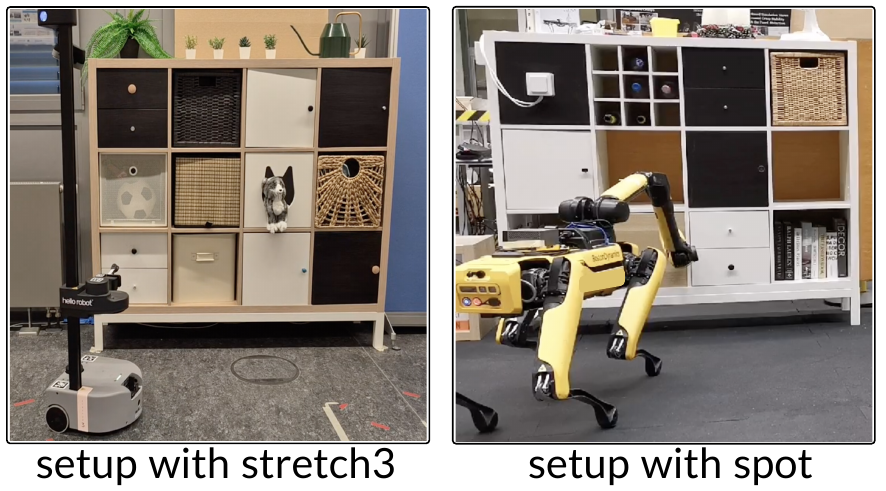}\vspace{-2mm}
    \caption{\textbf{Experimental setup.} Our KALLAX benchmark draws sample configs from the same distribution, which are then recreated in two different labs to evaluate the two different robotic systems.}
    \label{fig:exp_setup}
\end{figure}

Our experiments evaluate the benchmarking methodology itself rather than any single manipulation method, and are organized around three questions: (i)~\emph{reproducibility} -- whether seed-defined configurations can be reconstructed across different laboratories and robotic platforms under a common seed; (ii)~\emph{robustness} -- how strongly success rates vary across randomized configurations, and whether a capable system maintains consistent performance across them; and (iii)~\emph{comparability} -- whether the statistical framework of Section~\ref{sec:method} lets us draw controlled, confident comparisons between systems evaluated in different labs.

\subsection{Benchmarked Systems and Tasks}
We benchmark two heterogeneous robots in labs across two countries:
a Hello~Robot Stretch~3 and a Boston~Dynamics Spot with a
manipulator arm. Their differing locomotion, reach, and kinematics
support cross-embodiment evaluation.
Each platform uses a distinct but related autonomy stack.
Stretch~3 uses Stretch-Compose~\cite{stretch-compose}, which is
based on Spot-Compose~\cite{lemke2024spotcomposeframeworkopenvocabularyobject}.
Spot uses a stack based on
SpotLight~\cite{engelbracht2024spotlightroboticsceneunderstanding},
which in turn builds on Spot-Compose.
Both stacks provide environment reasoning, perception for handle
and object detection, and navigation and manipulation functionality
conditioned on location or natural-language goals.
Figure~\ref{fig:exp_setup} shows the evaluation protocol: a randomizer generates seeded configurations that both labs reconstruct from a shared specification before executing the common task set, recording outcomes per attempt. For the shared-seed experiments, a single researcher reconstructed the scenes in both laboratories. Thus, the experiments vary the laboratory, robot embodiment, and software stack while controlling for variability introduced by different reconstruction teams.


Within the KALLAX Benchmark, we evaluate two task types: door and drawer opening. For each, we consider attempts successful only if the robots open a door or drawer by at least \SI{10}{cm}. We analyze the two task types separately, using all doors or all drawers within a sampled configuration as the corresponding task set $M_s$.

Overall, we conduct 568 experimental trials. For doors, we run 62 trials over 6 seeds with Spot and 130 trials with 7 seeds for Stretch. For drawers, we run 88 trials over 6 seeds with Spot and 288 trials over 8 seeds with Stretch. As a reference for experimental effort, approximately 150 trials with Spot were conducted by a single experimenter over three days.

\subsection{Variance Analysis}

Table~\ref{tab:same_different_seeds} shows results for different subsets of evaluation seeds. Random subset estimates exhibit high seed variance, especially for Spot, confirming substantial variation in seed difficulty. Comparing the standard same-configuration protocol to random-seed subsets, we find that small-sample noise can even reverse the apparent robot ranking (e.g., Drawers, Spot subset~1 vs. Stretch subset~3). This highlights the need for statistical significance measures beyond raw success rates in stochastic benchmarks.


\subsection{Detectable performance margin}
\label{sec:exp-margin}

We invert the planning framework of Section~\ref{sec:method}: rather
than asking how many seeds are needed for a target margin, we ask, given
the seeds we have already evaluated, what is the smallest performance
difference between the two robots that we can resolve? Throughout this
analysis we adopt the pilot-study settings $\alpha=0.10$ and
$1-\beta=0.80$ -- yielding $z_{1-\alpha/2}+z_{1-\beta} \approx 2.49$ --
which trade a slightly higher Type~I rate for power at small samples.
We follow the per-task aggregation of Section~\ref{sec:method-setup}. 


For an unpaired comparison between Spot and Stretch, the standard error
of the difference in success rates for task type $t$ is $ SE_t$ and the smallest reliably detectable difference is
$\delta_{\min,t}$.
\begin{align*}
    \delta_{\min,t} = (z_{1-\alpha/2}+z_{1-\beta})\, SE_t \qquad SE_t = \sqrt{\frac{\hat V_{\mathrm{Spot},t}}{K_{\mathrm{Spot},t}}
                 + \frac{\hat V_{\mathrm{Stretch},t}}{K_{\mathrm{Stretch},t}}}
\end{align*}

Table~\ref{tab:experiment_detectable_margins} reports the observed gaps
and detectable margins. Both task types are distinguishable at the
chosen significance and power: Spot outperforms Stretch by $0.531$ on
doors against a detectable margin of $0.333$,  and
by $0.244$ on drawers against a margin of $0.232$. As a complementary uncertainty estimate, Welch $95\%$ confidence intervals for the Spot--Stretch difference are $23.4$--$82.8$ percentage points for doors and $4.2$--$44.6$ points for drawers, with Welch degrees of freedom $\nu=10.4$ and $\nu=12.9$ respectively. The drawer comparison thus sits close to the decision
boundary, and the framework's $1/\sqrt{K}$ scaling implies roughly
$4\times$ more seeds would be needed to halve the margin.
These results show that the framework can distinguish robotic systems evaluated in different laboratories and on different scenes, while also illustrating the cost of statistical resolution: smaller detectable margins require substantially more evaluation data.

\begin{table}[t]
    \centering
    \small
    \setlength{\tabcolsep}{3pt}
    \renewcommand{\arraystretch}{1.15}
    \caption{Testing on same vs. different shelf configurations. Averaged success rates vary substantially with the selected scene subset when few seeds are used.}
\begin{tabular}{lrrrr}
\toprule
Test Scenes & \multicolumn{2}{c}{Drawer Opening} & \multicolumn{2}{c}{Door Opening}\\
\cmidrule(lr){2-3} \cmidrule(lr){4-5}
& Spot & Stretch & Spot & Stretch\\
\midrule
same seeds & 92.7\% & 58.1\% & 86.3\% & 12.8\%\\
subset 1   & 65.0\% & 61.9\% & 66.7\% & 16.8\%\\
subset 2   & 100.0\% & 51.3\% & 75.0\% & 20.2\%\\
subset 3   & 87.5\% & 70.7\% & 100.0\% & 30.0\%\\
\bottomrule
\end{tabular}

    \label{tab:same_different_seeds}
\end{table}

\begin{table}[t]
\centering
\caption{Observed robot success rates and detectable margins. Margins are computed for \(\alpha=0.10\) and \(80\%\) power. Partial successes are counted as failures, and all configuration labels are included.}
\label{tab:experiment_detectable_margins}
\resizebox{\linewidth}{!}{%
\begin{tabular}{lrrrcrcr}
\toprule
& & & & \multicolumn{2}{c}{different random seeds} & \multicolumn{2}{c}{50/50 same and different seeds}\\
\cmidrule(lr){5-6} \cmidrule(lr){7-8}
Task & Spot & Stretch & difference &
detectable margin & result & detectable margin & result \\
\midrule
Door   & 0.7109 & 0.1802 & 0.5307 & 0.333 & distinguishable & 0.256 & distinguishable \\
Drawer & 0.8792 & 0.6352 & 0.2439 & 0.232 & distinguishable & 0.183 & distinguishable \\
\bottomrule
\end{tabular}}
\end{table}


\paragraph{Mixed shared-randomized comparison.}
As described in Section~\ref{sec:method-comparison}, a potential way to achieve lower detectable margins is to partially test robots on the same seeds. The right half of Table~\ref{tab:experiment_detectable_margins}
reports a mixed
design with \(p=0.5\), where half of the comparison is performed on shared
configurations and half on independently chosen configurations.
We observe that this reduces the detectable margin for both task types: from
\(33.3\) to \(25.6\) percentage points for doors and from \(23.2\) to \(18.3\)
percentage points for drawers. Mixed sharing therefore provides a practical compromise:
it reduces the required benchmarking effort while preserving evaluation on
independently sampled configurations. Importantly, both the fully independent and $p=0.5$ mixed designs lead to the same qualitative conclusion for the two systems in our experiments.


\section{Limitations}\label{sec:limitations}
A key limitation is that a considerable number of trials is required to obtain a statistically detectable performance margin. Moreover, because the same researcher reconstructed the shared-seed scenes in both laboratories, the present study does not quantify reconstruction variability that may arise when independent teams reconstruct the same configuration. Finally, the current evaluation relies on honest and well-calibrated reporting of success. Requiring proof with uploaded execution videos, and eventually automated VLM-based evaluation, could improve verifiability and standardization.

\section{Conclusion}
\label{sec:conclusion}
We introduce \emph{Stochastic Challenges}, a novel framework for comparing heterogeneous robotic systems on real-world tasks sampled from a shared distribution. Unlike fixed paired test sets, the framework supports independent, shared, and mixed configuration sampling, while explicitly accounting for execution noise and configuration difficulty. The KALLAX benchmark demonstrates the approach in mobile manipulation, and our two-lab study shows that statistically grounded cross-embodiment comparison is feasible without requiring identical robotic platforms or software stacks.

\clearpage
\acknowledgments {We thank Laurent Côté for support with the experiments at the University of Bonn and for valuable discussion on the statistical analysis, and Tim Engelbracht for assistance with the experiments at ETH Zurich. This work was partially supported by the Lamarr Institute for Machine Learning and Artificial Intelligence, the Robotics Institute Germany, and Google. Gülhan Şikaroğlu is supported by the YLSY Scholarship Program of the Republic of Türkiye Ministry of National Education under Law No.~1416.}

\bibliography{KALLAX-Benchmark}  

@misc{engelbracht2024spotlightroboticsceneunderstanding,
      title={SpotLight: Robotic Scene Understanding through Interaction and Affordance Detection}, 
      author={Tim Engelbracht and René Zurbrügg and Marc Pollefeys and Hermann Blum and Zuria Bauer},
      year={2024},
      eprint={2409.11870},
      archivePrefix={arXiv},
      primaryClass={cs.RO},
      url={https://arxiv.org/abs/2409.11870}, 
}

@misc{stretch-compose,
  author       = {{Robot Perception and Learning Lab}},
  title        = {Open-Vocabulary and Semantic-Aware Search and Retrieval of Objects using the Stretch Robot},
  howpublished = {\url{https://github.com/rpl-bonn/stretch-compose}},
  note         = {Accessed: 2026-02-26},
  year         = {2026}
}

@misc{lemke2024spotcomposeframeworkopenvocabularyobject,
      title={Spot-Compose: A Framework for Open-Vocabulary Object Retrieval and Drawer Manipulation in Point Clouds}, 
      author={Oliver Lemke and Zuria Bauer and René Zurbrügg and Marc Pollefeys and Francis Engelmann and Hermann Blum},
      year={2024},
      eprint={2404.12440},
      archivePrefix={arXiv},
      primaryClass={cs.RO},
      url={https://arxiv.org/abs/2404.12440}, 
}

@misc{Pum24,
      title={THE COLOSSEUM: A Benchmark for Evaluating Generalization for Robotic Manipulation}, 
      author={Wilbert Pumacay and Ishika Singh and Jiafei Duan and Ranjay Krishna and Jesse Thomason and Dieter Fox},
      year={2024},
      eprint={2402.08191},
      archivePrefix={arXiv},
      primaryClass={cs.RO},
      url={https://arxiv.org/abs/2402.08191}, 
}

@INPROCEEDINGS{Cal15,
  author={Calli, Berk and Singh, Arjun and Walsman, Aaron and Srinivasa, Siddhartha and Abbeel, Pieter and Dollar, Aaron M.},
  booktitle={2015 International Conference on Advanced Robotics (ICAR)}, 
  title={The YCB object and Model set: Towards common benchmarks for manipulation research}, 
  year={2015},
  volume={},
  number={},
  pages={510-517},
  doi={10.1109/ICAR.2015.7251504}}

@misc{Atr25,
      title={RoboArena: Distributed Real-World Evaluation of Generalist Robot Policies}, 
      author={Pranav Atreya and Karl Pertsch and Tony Lee and Moo Jin Kim and Arhan Jain and Artur Kuramshin and Clemens Eppner and Cyrus Neary and Edward Hu and Fabio Ramos and Jonathan Tremblay and Kanav Arora and Kirsty Ellis and Luca Macesanu and Marcel Torne Villasevil and Matthew Leonard and Meedeum Cho and Ozgur Aslan and Shivin Dass and Jie Wang and William Reger and Xingfang Yuan and Xuning Yang and Abhishek Gupta and Dinesh Jayaraman and Glen Berseth and Kostas Daniilidis and Roberto Martin-Martin and Youngwoon Lee and Percy Liang and Chelsea Finn and Sergey Levine},
      year={2025},
      eprint={2506.18123},
      archivePrefix={arXiv},
      primaryClass={cs.RO},
      url={https://arxiv.org/abs/2506.18123}, 
}

@misc{Che25b,
      title={ManipulationNet: An Infrastructure for Benchmarking Real-World Robot Manipulation with Physical Skill Challenges and Embodied Multimodal Reasoning}, 
      author={Yiting Chen and Kenneth Kimble and Edward H. Adelson and Tamim Asfour and Podshara Chanrungmaneekul and Sachin Chitta and Yash Chitambar and Ziyang Chen and Ken Goldberg and Danica Kragic and Hui Li and Xiang Li and Yunzhu Li and Aaron Prather and Nancy Pollard and Maximo A. Roa-Garzon and Robert Seney and Shuo Sha and Shihefeng Wang and Yu Xiang and Kaifeng Zhang and Yuke Zhu and Kaiyu Hang},
      year={2026},
      eprint={2603.04363},
      archivePrefix={arXiv},
      primaryClass={cs.RO},
      url={https://arxiv.org/abs/2603.04363}, 
}

@ARTICLE{Jam20,
  author={James, Stephen and Ma, Zicong and Arrojo, David Rovick and Davison, Andrew J.},
  journal={IEEE Robotics and Automation Letters}, 
  title={RLBench: The Robot Learning Benchmark \& Learning Environment}, 
  year={2020},
  volume={5},
  number={2},
  pages={3019-3026},
  doi={10.1109/LRA.2020.2974707}}

@INPROCEEDINGS{Gar25,
  author={Garcia, Ricardo and Chen, Shizhe and Schmid, Cordelia},
  booktitle={2025 IEEE International Conference on Robotics and Automation (ICRA)}, 
  title={Towards Generalizable Vision-Language Robotic Manipulation: A Benchmark and LLM-Guided 3D Policy}, 
  year={2025},
  volume={},
  number={},
  pages={8996-9002},
  doi={10.1109/ICRA55743.2025.11127315}}

@misc{Zhao25,
      title={ManipBench: Benchmarking Vision-Language Models for Low-Level Robot Manipulation}, 
      author={Enyu Zhao and Vedant Raval and Hejia Zhang and Jiageng Mao and Zeyu Shangguan and Stefanos Nikolaidis and Yue Wang and Daniel Seita},
      year={2025},
      eprint={2505.09698},
      archivePrefix={arXiv},
      primaryClass={cs.RO},
      url={https://arxiv.org/abs/2505.09698}, 
}

@misc{Zha25c,
      title={VLABench: A Large-Scale Benchmark for Language-Conditioned Robotics Manipulation with Long-Horizon Reasoning Tasks}, 
      author={Shiduo Zhang and Zhe Xu and Peiju Liu and Xiaopeng Yu and Yuan Li and Qinghui Gao and Zhaoye Fei and Zhangyue Yin and Zuxuan Wu and Yu-Gang Jiang and Xipeng Qiu},
      year={2024},
      eprint={2412.18194},
      archivePrefix={arXiv},
      primaryClass={cs.RO},
      url={https://arxiv.org/abs/2412.18194}, 
}

@misc{Nas24,
      title={RoboCasa: Large-Scale Simulation of Everyday Tasks for Generalist Robots}, 
      author={Soroush Nasiriany and Abhiram Maddukuri and Lance Zhang and Adeet Parikh and Aaron Lo and Abhishek Joshi and Ajay Mandlekar and Yuke Zhu},
      year={2024},
      eprint={2406.02523},
      archivePrefix={arXiv},
      primaryClass={cs.RO},
      url={https://arxiv.org/abs/2406.02523}, 
}

@misc{Li22,
      title={BEHAVIOR-1K: A Human-Centered, Embodied AI Benchmark with 1,000 Everyday Activities and Realistic Simulation}, 
      author={Chengshu Li and Ruohan Zhang and Josiah Wong and Cem Gokmen and Sanjana Srivastava and Roberto Martín-Martín and Chen Wang and Gabrael Levine and Wensi Ai and Benjamin Martinez and Hang Yin and Michael Lingelbach and Minjune Hwang and Ayano Hiranaka and Sujay Garlanka and Arman Aydin and Sharon Lee and Jiankai Sun and Mona Anvari and Manasi Sharma and Dhruva Bansal and Samuel Hunter and Kyu-Young Kim and Alan Lou and Caleb R Matthews and Ivan Villa-Renteria and Jerry Huayang Tang and Claire Tang and Fei Xia and Yunzhu Li and Silvio Savarese and Hyowon Gweon and C. Karen Liu and Jiajun Wu and Li Fei-Fei},
      year={2024},
      eprint={2403.09227},
      archivePrefix={arXiv},
      primaryClass={cs.RO},
      url={https://arxiv.org/abs/2403.09227}, 
}

@misc{Luo25,
      title={FMB: a Functional Manipulation Benchmark for Generalizable Robotic Learning}, 
      author={Jianlan Luo and Charles Xu and Fangchen Liu and Liam Tan and Zipeng Lin and Jeffrey Wu and Pieter Abbeel and Sergey Levine},
      year={2024},
      eprint={2401.08553},
      archivePrefix={arXiv},
      primaryClass={cs.RO},
      url={https://arxiv.org/abs/2401.08553}, 
}

@misc{Heo25,
      title={FurnitureBench: Reproducible Real-World Benchmark for Long-Horizon Complex Manipulation}, 
      author={Minho Heo and Youngwoon Lee and Doohyun Lee and Joseph J. Lim},
      year={2023},
      eprint={2305.12821},
      archivePrefix={arXiv},
      primaryClass={cs.RO},
      url={https://arxiv.org/abs/2305.12821}, 
}

@inproceedings{arnold,
  author = {Gong, Ran and Huang, Jiangyong and Zhao, Yizhou and Geng, Haoran and Gao, Xiaofeng and Wu, Qingyang and Ai, Wensi and Zhou, Ziheng and Terzopoulos, Demetri and Zhu, Song-Chun and Jia, Baoxiong and Huang, Siyuan},
  title = {{ARNOLD}: A Benchmark for Language-Grounded Task Learning with Continuous States in Realistic 3D Scenes},
  booktitle = {Proceedings of the IEEE/CVF International Conference on Computer Vision (ICCV)},
  year = {2023},
  pages = {20483--20495},
  doi = {10.1109/ICCV51070.2023.01873},
  url = {https://arxiv.org/abs/2304.04321}
}

@article{calvin,
  title={CALVIN: A Benchmark for Language-Conditioned Policy Learning for Long-Horizon Robot Manipulation Tasks},
  author={Mees, Oier and Hermann, Lukas and Rosete-Victoria, Erick and Burgard, Wolfram},
  journal={IEEE Robotics and Automation Letters (RA-L)},
  volume={7},
  number={3},
  pages={6323--6330},
  year={2022},
  publisher={IEEE},
  doi={10.1109/LRA.2022.3173719},
  url={https://arxiv.org/abs/2112.03227}
}

@misc{libero,
      title={LIBERO: Benchmarking Knowledge Transfer for Lifelong Robot Learning}, 
      author={Bo Liu and Yifeng Zhu and Chongkai Gao and Yihao Feng and Qiang Liu and Yuke Zhu and Peter Stone},
      year={2023},
      eprint={2306.03310},
      archivePrefix={arXiv},
      primaryClass={cs.AI},
      url={https://arxiv.org/abs/2306.03310}, 
}

@misc{scenereplica,
      title={SCENEREPLICA: Benchmarking Real-World Robot Manipulation by Creating Replicable Scenes}, 
      author={Ninad Khargonkar and Sai Haneesh Allu and Yangxiao Lu and Jishnu Jaykumar P and Balakrishnan Prabhakaran and Yu Xiang},
      year={2024},
      eprint={2306.15620},
      archivePrefix={arXiv},
      primaryClass={cs.RO},
      url={https://arxiv.org/abs/2306.15620}, 
}

@misc{robodojo,
      title={RoboDojo: A Unified Sim-and-Real Benchmark for Comprehensive Evaluation of Generalist Robot Manipulation Policies}, 
      author={Tianxing Chen and Yue Chen and Zixuan Li and Junyuan Tang and Kailun Su and Weijie Wan and Baijun Chen and Haoran Lu and Haowen Yan and Honghao Su and Zhiyang Dou and Kaixuan Wang and Dandan Zhang and Yunze Liu and Yan Qin and Qiwei Liang and Qiwei Wu and Zijian Lin and Wenwei Lin and Yuran Wang and Minghua He and Tianshu Wu and Ruihai Wu and Jingquan Zhou and Kai-Chong Lei and Haibao Yu and Yuanfeng Ji and Weiyang Jin and Guanyu Lin and Xiaofan Li and Qi Xiong and Renjing Xu and Zhongyu Li and Wenhao Chai and Enze Xie and Ziwei Wang and Yao Mu and Hao Dong and Wojciech Matusik and Mingyu Ding and Wenbo Ding and Ping Luo and Masayoshi Tomizuka},
      year={2026},
      eprint={2607.04434},
      archivePrefix={arXiv},
      primaryClass={cs.RO},
      url={https://arxiv.org/abs/2607.04434}
}

@article{graspa,
  author   = {Bottarel, F. and Altobelli, A. and Pattacini, U. and Natale, L.},
  journal  = {IEEE Robotics \& Automation Magazine},
  title    = {GRASPA-fying the Panda: Easily Deployable, Fully Reproducible Benchmarking of Grasp Planning Algorithms},
  year     = {2024},
  volume   = {31},
  number   = {4},
  pages    = {21--31},
  doi      = {10.1109/MRA.2023.3256272},
}

@article{Morgan20,
  author  = {Morgan, Andrew S. and Hang, Kaiyu and Bircher, Walter G. and Alladkani, Fadi M. and Gandhi, Abhinav and Calli, Berk and Dollar, Aaron M.},
  journal = {IEEE Robotics and Automation Letters},
  title   = {Benchmarking Cluttered Robot Pick-and-Place Manipulation With the Box and Blocks Test},
  year    = {2020},
  volume  = {5},
  number  = {2},
  pages   = {454--461},
  doi     = {10.1109/LRA.2019.2961053}
}

@article{Bauer2021Real,
  author  = {Bauer, Stefan and Widmaier, Felix and Wüthrich, Manuel and Buchholz, Annika and Stark, Sebastian and Goyal, Anirudh and Steinbrenner, Thomas and Akpo, Joel and Joshi, Shruti and Berenz, Vincent and Agrawal, Vaibhav and Funk, Niklas and De Jesus, Julen Urain and Peters, Jan and Watson, Joe and Chen, Claire and Srinivasan, Krishnan and Zhang, Junwu and Zhang, Jeffrey and Walter, Matthew R. and Madan, Rishabh and Schaff, Charles and Maeda, Takahiro and Yoneda, Takuma and Yarats, Denis},
  journal = {arXiv preprint arXiv:2109.10957},
  title   = {Real Robot Challenge: A Robotics Competition in the Cloud},
  year    = {2021},
  doi     = {10.48550/arXiv.2109.10957}
}

@inproceedings{RB2,
  author    = {Dasari, Sudeep and Wang, Jianren and Hong, Joyce and Bahl, Shikhar and Lin, Yixin and Wang, Austin and Thankaraj, Abitha and Chahal, Karanbir and Calli, Berk and Gupta, Saurabh and Held, David and Pinto, Lerrel and Pathak, Deepak and Kumar, Vikash and Gupta, Abhinav},
  booktitle = {Thirty-fifth Conference on Neural Information Processing Systems Datasets and Benchmarks Track},
  title     = {RB2: Robotic Manipulation Benchmarking with a Twist},
  year      = {2021}
}

@inproceedings{NIST,
  author    = {Lian, Wenzhao and Kelch, Tim and Holz, Dirk and Norton, Adam and Schaal, Stefan},
  booktitle = {2021 IEEE/RSJ International Conference on Intelligent Robots and Systems (IROS)},
  title     = {Benchmarking Off-The-Shelf Solutions to Robotic Assembly Tasks},
  year      = {2021},
  pages     = {1046--1053},
  doi       = {10.1109/IROS51168.2021.9636586}
}

@article{OCRTOC,
  author  = {Liu, Ziyuan and Liu, Wei and Qin, Yuzhe and Xiang, Fanbo and Gou, Minghao and Xin, Songyan and Roa, Maximo A. and Calli, Berk and Su, Hao and Sun, Yu and Tan, Ping},
  journal = {IEEE Robotics and Automation Letters},
  title   = {OCRTOC: A Cloud-Based Competition and Benchmark for Robotic Grasping and Manipulation},
  year    = {2022},
  volume  = {7},
  number  = {1},
  pages   = {486--493},
  doi     = {10.1109/LRA.2021.3129136}
}

@inproceedings{TOTO,
  author    = {Zhou, Gaoyue and Dean, Victoria and Srirama, Mohan Kumar and Rajeswaran, Aravind and Pari, Jyothish and Hatch, Kyle and Jain, Aryan and Yu, Tianhe and Abbeel, Pieter and Pinto, Lerrel and Finn, Chelsea and Gupta, Abhinav},
  booktitle = {2023 IEEE International Conference on Robotics and Automation (ICRA)},
  title     = {Train Offline, Test Online: A Real Robot Learning Benchmark},
  year      = {2023},
  pages     = {9197--9203},
  doi       = {10.1109/ICRA48891.2023.10160594}
}
\clearpage
\appendix
\section{Estimators}\label{app:estimators}

This appendix derives the elementary properties of the per-task,
seed-level, and benchmark-mean estimators introduced in
Section~\ref{sec:method-setup}.

\paragraph{Per-trial model.}
For a fixed robot $R$, seed $s$, and task $i$, repeated trials are
independent  trials with unknown success probability $p_{Rsi}$:
\[
    Y_{Rsij} \overset{\mathrm{iid}}{\sim} \mathrm{Bernoulli}(p_{Rsi}),
\]
so that $\mathbb{E}[Y_{Rsij}] = p_{Rsi}$ and $\mathrm{Var}(Y_{Rsij}) = p_{Rsi}(1-p_{Rsi})$.

\paragraph{Per-task estimator.}
For $J$ repeated trials of task $i$ in seed $s$,
\[
    \hat p_{Rsi} = \frac{1}{J} \sum_{j=1}^{J} Y_{Rsij}.
\]
Conditional on $p_{Rsi}$,
\[
    \mathbb{E}[\hat p_{Rsi} \mid p_{Rsi}] = p_{Rsi},
    \qquad
    \mathrm{Var}(\hat p_{Rsi} \mid p_{Rsi}) = \frac{p_{Rsi}(1-p_{Rsi})}{J},
\]
so $\hat p_{Rsi}$ is unbiased for $p_{Rsi}$ and its variance shrinks as
$1/J$.

\paragraph{Seed-level estimator.}
The seed-level score averages task estimates uniformly,
\[
    \hat\theta_{Rs} = \frac{1}{M_s} \sum_{i=1}^{M_s} \hat p_{Rsi},
\]
which is conditionally unbiased for the true seed-level performance
$\theta_{Rs} = \tfrac{1}{M_s}\sum_i p_{Rsi}$:
\[
    \mathbb{E}\!\left[\hat\theta_{Rs} \mid \{p_{Rsi}\}\right]
    = \frac{1}{M_s}\sum_i p_{Rsi}
    = \theta_{Rs}.
\]
The uniform average over tasks, rather than over raw trials, prevents
tasks retried more often from exerting disproportionate influence on
$\hat\theta_{Rs}$.

\paragraph{Benchmark mean.}
Over $K$ independent seeds drawn from the benchmark distribution,
\[
    \hat\mu_R = \frac{1}{K}\sum_{s=1}^{K} \hat\theta_{Rs}.
\]
Taking expectation over both trial noise and seed sampling,
\[
    \mathbb{E}[\hat\mu_R]
    = \mathbb{E}_S[\theta_{Rs}]
    = \mu_R,
\]
so $\hat\mu_R$ is an unbiased estimator of the benchmark objective
$\mu_R$.

\section{Variance}\label{app:variance}
We derive the planning variance
$V_R(J) = \sigma^2_{\mathrm{seed},R} + B_R/J$ used in
Section~\ref{sec:method-variance}.

\paragraph{Within-seed variance.}
Conditional on a seed $s$, treating per-task estimates
$\{\hat p_{Rsi}\}$ as conditionally independent,
\[
    \mathrm{Var}(\hat\theta_{Rs} \mid s)
    = \frac{1}{M_s^2}\sum_{i=1}^{M_s}
        \mathrm{Var}(\hat p_{Rsi}\mid p_{Rsi})
    = \frac{1}{M_s^2}\sum_{i=1}^{M_s} \frac{p_{Rsi}(1-p_{Rsi})}{J}.
\]
Defining the per-seed Bernoulli-variance aggregate
\[
    B_{Rs} = \frac{1}{M_s^2}\sum_{i=1}^{M_s} p_{Rsi}(1-p_{Rsi}),
\]
this becomes $\mathrm{Var}(\hat\theta_{Rs} \mid s) \approx B_{Rs}/J$.

\paragraph{Between-seed variance.}
Configurations differ in difficulty, so the true seed-level performance
varies across seeds:
\[
    \sigma^2_{\mathrm{seed},R} = \mathrm{Var}_S(\theta_{Rs}).
\]

\paragraph{Total variance via law of total variance.}
Treating $\hat\theta_{Rs}$ as random over the joint distribution of
seeds and trial noise, the law of total variance gives
\[
    \mathrm{Var}(\hat\theta_{Rs})
    = \mathrm{Var}\!\left(\mathbb{E}[\hat\theta_{Rs}\mid S]\right)
      + \mathbb{E}\!\left[\mathrm{Var}(\hat\theta_{Rs}\mid S)\right].
\]
Substituting $\mathbb{E}[\hat\theta_{Rs}\mid S=s] = \theta_{Rs}$
(Appendix~\ref{app:estimators}) and
$\mathrm{Var}(\hat\theta_{Rs}\mid S=s) \approx B_{Rs}/J$,
\[
    \mathrm{Var}(\hat\theta_{Rs})
    = \mathrm{Var}_S(\theta_{Rs}) + \mathbb{E}_S\!\left[\frac{B_{Rs}}{J}\right]
    = \sigma^2_{\mathrm{seed},R} + \frac{B_R}{J},
\]
where $B_R = \mathbb{E}_S[B_{Rs}]$. We denote this combined quantity the
\emph{planning variance}
$V_R(J) = \sigma^2_{\mathrm{seed},R} + B_R/J$.

\paragraph{Benchmark mean variance.}
Under independent seed draws, averaging over $K$ seeds gives
\[
    \mathrm{Var}(\hat\mu_R) = \frac{V_R(J)}{K}
    = \frac{\sigma^2_{\mathrm{seed},R} + B_R/J}{K}.
\]

\section{Comparison}\label{app:comparison}

We derive the variance of the difference estimator
$\hat\Delta = \hat\mu_A - \hat\mu_B$ under the unpaired, fully paired,
and mixed designs of
Sections~\ref{sec:method-samplesize}--\ref{sec:method-comparison}.

\paragraph{Unpaired comparison.}
If robots $A$ and $B$ are evaluated on independent sets of $K$ seeds
each, $\hat\mu_A$ and $\hat\mu_B$ are independent, and
\[
    \mathrm{Var}(\hat\Delta)
    = \mathrm{Var}(\hat\mu_A) + \mathrm{Var}(\hat\mu_B)
    = \frac{V_A(J) + V_B(J)}{K}
    \approx \frac{2V(J)}{K},
\]
where the last step assumes comparable planning variance
$V_A(J)\approx V_B(J)\approx V(J)$. Substituting into the
normal-approximation power requirement
$\delta = (z_{1-\alpha/2}+z_{1-\beta})\sqrt{\mathrm{Var}(\hat\Delta)}$
and solving for $K$ yields the sample-size
formula~\eqref{eq:samplesize} of
Section~\ref{sec:method-samplesize}.

\paragraph{Paired comparison.}
If both robots are evaluated on the same $K$ seeds, we model each
seed-level score as
\[
    \hat\theta_{As} = \mu_A + a_s + \varepsilon_{As},
    \qquad
    \hat\theta_{Bs} = \mu_B + a_s + \varepsilon_{Bs},
\]
with a common seed-difficulty effect $a_s$ satisfying
$\mathrm{Var}(a_s) = \sigma^2_{\mathrm{seed}}$ and residual within-seed
trial noise $\varepsilon_{Rs}$ with
$\mathrm{Var}(\varepsilon_{Rs}) \approx B/J$. The per-seed difference
\[
    D_s = \hat\theta_{As} - \hat\theta_{Bs}
        = \Delta + \varepsilon_{As} - \varepsilon_{Bs}
\]
cancels $a_s$ exactly. Assuming the residuals are independent across
robots given the seed,
\[
    \mathrm{Var}(D_s) \approx \frac{2B}{J},
    \qquad
    \mathrm{Var}(\bar D) \approx \frac{2B/J}{K},
\]
removing the between-seed variance term entirely.

\paragraph{Mixed design.}
For a fraction $p \in [0,1]$ of seeds shared between the two robots,
the variance of $\hat\Delta$ interpolates between the unpaired and
fully paired endpoints. Treating shared seeds as paired and unshared
seeds as independent, an additive approximation gives
\[
    \mathrm{Var}(\hat\Delta)
    \approx \frac{2\bigl[(1-p)\,\sigma^2_{\mathrm{seed}} + B/J\bigr]}{K},
\]
which recovers $2V(J)/K$ at $p=0$ and $2B/J\,/\,K$ at $p=1$.
Substituting into the power requirement gives the mixed-design
sample-size formula of Section~\ref{sec:method-comparison}.

\section{Extensions}\label{app:extensions}

We describe three refinements of the analysis used in the experiments
as stratified reports, a model-based robustness check, and the basis
for equivalence claims.

\paragraph{Task-type stratification.}
Different interaction types differ systematically in difficulty -- a
recessed bottom drawer is not comparable to an open shelf at chest
height. The variance components and benchmark scores can therefore be
estimated separately for each task class
\[
    T \in \{\text{door},\ \text{drawer-top},\ \text{drawer-bottom},\ \ldots\}.
\]
Restricting the per-task estimators of Appendix~\ref{app:estimators} to
tasks of class $T$ yields stratified estimates $\mu_{R,T}$,
$\sigma^2_{\mathrm{seed},T}$, and $B_T$. These allow different task
classes to have different mean success rates and variance components,
and prevent unbalanced trial mixes from biasing the overall comparison.

\paragraph{Logistic mixed-effects model.}
The variance decomposition of Appendix~\ref{app:variance} relies on a
normal approximation to averaged Bernoulli outcomes. As a more exact
alternative, we fit a logistic mixed-effects model directly to the
trial-level binary outcomes:
\[
    Y_{Rsij} \sim \mathrm{Bernoulli}(p_{Rsij}),
\]
\[
    \mathrm{logit}(p_{Rsij})
    = \mu + \beta_R + \beta_T + (\beta_R \beta_T) + a_s + b_{si},
\]
with fixed robot ($\beta_R$) and task-type ($\beta_T$) effects, their
interaction $(\beta_R\beta_T)$, a random seed effect $a_s$, and a
random task-within-seed effect $b_{si}$. The model is used as a
robustness check on the headline comparison.

\paragraph{Equivalence testing.}
Failure to detect a difference at significance level $\alpha$ is not
evidence that two robots are equivalent. To claim practical equivalence
we specify in advance an equivalence margin $\delta_{\mathrm{eq}} > 0$
and conclude that $A$ and $B$ are practically equivalent if a
$100(1-\alpha)\%$ confidence interval for $\Delta = \mu_A - \mu_B$ lies
entirely within $[-\delta_{\mathrm{eq}},\, \delta_{\mathrm{eq}}]$. The
choice of $\delta_{\mathrm{eq}}$ reflects the smallest performance
difference considered practically meaningful for the benchmark.
\end{document}